\documentclass[10pt,journal,compsoc]{IEEEtran}

\usepackage{hyperref}
\usepackage{booktabs}
\usepackage{multirow}
\usepackage{color}
\usepackage[table,xcdraw]{xcolor}
\usepackage{stix,bbding,pifont,utfsym,fontawesome}
\usepackage{amsmath}

\makeatletter
\let\MYcaption\@makecaption
\makeatother

\usepackage[font=footnotesize]{subcaption}

\makeatletter
\let\@makecaption\MYcaption
\makeatother

\ifCLASSOPTIONcompsoc
  \usepackage[nocompress]{cite}
\else
  \usepackage{cite}
\fi

\ifCLASSINFOpdf
\else
\fi

\def\dataset{OWSC}
\def\datasetb{OWSC }

\usepackage{xr}
\begin{document}
\title{A Dataset and Framework for Learning State-invariant Object Representations}
\author{Rohan Sarkar, Avinash Kak\\
School of Electrical and Computer Engineering, Purdue University, USA}
\markboth{Sarkar \MakeLowercase{\textit{et al.}}: A Dataset and Framework for Learning State-invariant Object Representations}%
{Shell \MakeLowercase{\textit{et al.}}: Learning State-Invariant Representations of Objects}

\IEEEtitleabstractindextext{%
\begin{abstract}
We add one more invariance --- the state invariance --- to the more
commonly used other invariances for learning object representations
for recognition and retrieval.  By state invariance, we mean robust
with respect to changes in the structural form of the objects, such as
when an umbrella is folded, or when an item of clothing is tossed on the
floor.  Since humans generally have no difficulty recognizing
objects despite such state changes, we are naturally faced with the
question of whether it is possible to devise a neural architecture with
similar abilities. To that end, we present a novel dataset,
ObjectsWithStateChange, which captures state and pose variations in the
object images recorded from arbitrary viewpoints.  We believe that
this dataset will facilitate research in fine-grained object
recognition and retrieval of 3D objects that are capable of state
changes.  The goal of such research would be to train models capable
of learning discriminative object embeddings that remain invariant to state changes
while also staying invariant to transformations induced by changes in
 viewpoint, pose, illumination, etc. A major challenge in this regard is that instances of different objects (both within and across different categories) under various state changes may share similar visual characteristics and therefore may be close to one another in the learned embedding space, which would make it more difficult to discriminate between them. 
To address this, we propose a curriculum learning strategy that leverages the similarity relationships in the learned embedding space after each epoch to guide the training process. In accordance with the curriculum learning principles, during the training phase we progressively 
select object pairs with smaller inter-object distances in the learned embedding space, thereby gradually sampling harder-to-distinguish examples of visually similar objects, both within and across different categories. 
Our ablation related to the role played by curriculum learning indicates an improvement in object recognition accuracy of 7.9\% and retrieval mAP of 9.2\% over the state-of-the-art. We believe that this strategy enhances the
model's ability to capture discriminative features for fine-grained
tasks that involve objects with state changes, leading
to performance improvements on object-level tasks not only on the new
dataset we present, but also on three other challenging multi-view
datasets such as ModelNet40, ObjectPI, and FG3D.
\end{abstract}

}

\maketitle

\IEEEdisplaynontitleabstractindextext
\IEEEpeerreviewmaketitle

\begin{figure}[t]
    \centering
    \includegraphics[width=0.5\textwidth]{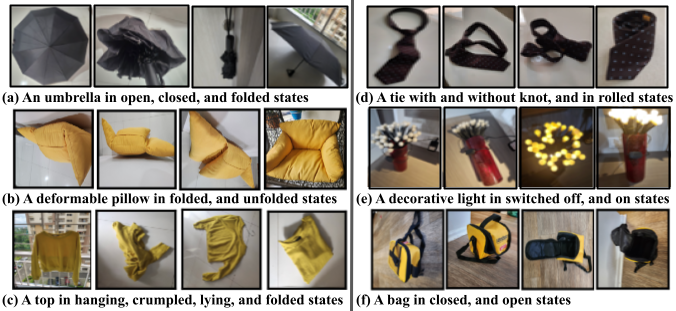}
    \caption{In addition to pose, many commonly occurring objects
      also exhibit significant changes in their appearance when their
      state changes. This figure displays several examples of such
      objects in our ObjectsWithStateChange (OWSC) dataset.}
    \label{fig:transformations}
\end{figure}

\IEEEraisesectionheading{\section{Introduction}}
\label{sec:intro}
\IEEEPARstart{T}{he} fact that the appearance of an object depends significantly on its
pose vis-a-vis that of the camera has received much attention in the
research literature.  However, there exists a more confounding cause
for major changes in object appearance --- the change in the state of
the object.  Consider, for example, objects like umbrellas, books on a table that may be
open or closed,
furniture items that are likely to be folded up when not in use, items
of clothing and linen in different forms depending on how they are stowed, etc.  Fig. \ref{fig:transformations} shows some examples of such
objects from our ObjectsWithStateChange dataset.

We posit that with regard to the objects we encounter in our daily
lives, it is just as likely as not that you will encounter objects
whose appearance depends significantly on what state we find them in.
That makes important the following question: Can modern computer
vision algorithms effectively recognize objects despite the changes in
their state?  That is, in addition to achieving pose and viewpoint
invariances, is it possible to also achieve invariance with respect to
state changes?  We believe that the work we present is a
step in the direction of answering those questions.

\begin{table}[t]
\caption{Comparison of the new ObjectsWithStateChange (OWSC) dataset with the existing multi-view object classification datasets. }
\setlength{\tabcolsep}{1pt}
\tiny
\begin{tabular}{@{}lclcccccccccc@{}}
\toprule
\multirow{2}{*}{Datasets} & \multirow{2}{*}{Year} & \multirow{2}{*}{\begin{tabular}[c]{@{}l@{}}Type of\\ Data\end{tabular}} & \multirow{2}{*}{\# Catg.} & \multicolumn{2}{c}{\# Objects} & \multirow{2}{*}{\begin{tabular}[c]{@{}c@{}}Real-\\world\\ objects\end{tabular}} & \multirow{2}{*}{\begin{tabular}[c]{@{}c@{}}In-the-wild\\ environment\\(Complex b/g)\end{tabular}} & \multirow{2}{*}{\begin{tabular}[c]{@{}c@{}}Fine-\\ grained\end{tabular}} & \multirow{2}{*}{\begin{tabular}[c]{@{}c@{}}Arbitrary\\ views\end{tabular}} & \multirow{2}{*}{\begin{tabular}[c]{@{}c@{}}State \\ changes\end{tabular}} & \multirow{2}{*}{\begin{tabular}[c]{@{}c@{}}Train and\\ Test split\end{tabular}} \\ \cmidrule{5-6}
 &  &  &  & Train & Test &  &  &  &  &  &  &  \\ \midrule
ModelNet40 \cite{ModelNet40} & 2015 & CAD & 40 & 3183 & 800 & {\XSolid} & {{\XSolid}} & {\XSolid} & { \XSolid}  & { \XSolid} & { { \CheckmarkBold}} \\
FG3D \cite{FG3D} & 2021 & CAD & 66 & 21575 & 3977 & { \XSolid} & { \XSolid} & { \CheckmarkBold} & { \XSolid} & { \XSolid} & { \CheckmarkBold} \\ \midrule
ETH-80 \cite{ETH80} & 2003 & RGB & 8 & 80 & 80 & { \CheckmarkBold} & { \XSolid} & { \XSolid} & { \XSolid} & { \XSolid} & { \CheckmarkBold} \\
RGB-D \cite{RGBD} & 2011 & RGB+Depth & 51 & 300 & 300 & { \CheckmarkBold} & { \XSolid} & { \XSolid} & { \XSolid} & { \XSolid} & { \CheckmarkBold} \\
MIRO \cite{RotationNet2018} & 2018 & RGB & 12 & 120 & 120 & { \CheckmarkBold} & { \XSolid} & { \XSolid} & { \XSolid} & { \XSolid} & { \CheckmarkBold} \\
Toybox \cite{Toybox} & 2018 & Video & 3 & 12 & 12 & { \CheckmarkBold} & { \XSolid} & { \XSolid} & { \CheckmarkBold}  & { \XSolid} & { \CheckmarkBold} \\
ObjectPI \cite{PIE2019} & 2019 & RGB & 25 & 382 & 98 & { \CheckmarkBold} & { \CheckmarkBold} & { \XSolid} & { \XSolid}  & { \XSolid} & { \CheckmarkBold} \\
ObjectNet \cite{ObjectNet} & 2019 & RGB & 313 & - & - & { \CheckmarkBold} & { \CheckmarkBold} & { \XSolid} & { \CheckmarkBold}  & { \XSolid} & { \XSolid} \\
RPC \cite{RPC2019} & 2022 & RGB & 17 & 200 & 200 & { \CheckmarkBold} & { \XSolid} & { \CheckmarkBold} & { \XSolid} & { \XSolid} & { \CheckmarkBold} \\
MVP-N \cite{MVP-N} & 2022 & RGB & - & 44 & 44 & { \CheckmarkBold} & { \XSolid} & { \CheckmarkBold} & { \XSolid}  & { \XSolid} & { \CheckmarkBold} \\ \midrule
\textbf{\begin{tabular}[c]{@{}l@{}}ObjectsWith\\StateChange (Ours)\end{tabular}} & 2025 & RGB+Text & 21 & 331 & 331 & { \CheckmarkBold} & { \CheckmarkBold} & { \CheckmarkBold} & { \CheckmarkBold}  & { \CheckmarkBold} & { \CheckmarkBold} \\ \bottomrule
\end{tabular}
\label{tbl:comp_ObjectTransform}
\end{table}
Obviously, any exploration of the answers to such questions
must begin with the creation of a dataset that captures not only pose
and viewpoint appearance changes, but also state changes.  To that
end, a primary goal of this paper is the presentation of a new
dataset we have created --- the ObjectsWithStateChange dataset ({\bf \dataset}).  We
believe that this dataset nicely augments the other well-known
datasets --- ModelNet40, ObjectPI, and FG3D --- that are only meant to
train and test for invariances with respect to pose and viewpoint.  We
believe that the \datasetb dataset would facilitate
creating image representations that would also be invariant to highly
nonlinear shape deformations caused by state changes in addition to
being invariant to changes in viewpoint, pose, illumination, etc.

To highlight the usefulness of the \datasetb dataset, we first evaluate several state-of-the-art pose-invariant methods \cite{PIE2019}, \cite{PiRO}, and conclude that these methods struggle with object recognition and retrieval tasks when objects undergo state changes.   
Subsequently, we propose an extension to the dual-encoder model
\cite{PiRO} that learns discriminative object and category
representations simultaneously by comparing images under different transformations of a pair of objects that are randomly sampled from the same category. However, 
it doesn’t allow the model to distinguish subtle differences between similar items, like half-sleeve and full-sleeve tops within the same category, or visually similar items like towels and blankets in different categories. The problem becomes even more challenging under various state changes, such as when clothing items are folded or piled.
In order to accommodate such changes in object state, we extend
\cite{PiRO} by proposing a curriculum learning strategy that uses the
inter-object distances in the learned embedding space after each epoch
to sample pairs of objects that have similar visual characteristics and are harder to distinguish in the presence of state changes.

Following curriculum learning principles, we progressively select object pairs with closer inter-object distances in the learned embedding space as training advances, thus gradually introducing harder-to-distinguish examples of visually similar objects within and across categories. Specifically, this strategy involves sampling images of similar objects with various state changes from the same category (by retrieving neighboring objects within the category in the learned embedding space) and from different categories (by subdividing the embedding space further as training progresses and sampling nearby objects within each partition). Although this method introduces the added complexity of finding a neighboring object for each object to form pairs, compared to simple random sampling within a category \cite{PiRO}, it maintains sub-linear time complexity relative to the number of objects per category using fast approximate nearest neighbor techniques (see Appendix \ref{sec:time_complexity}). This approach promotes cross-category and within-category comparisons, encouraging the model to differentiate between challenging examples and enhancing its ability to capture discriminative features for fine-grained tasks while being invariant to state and pose changes. This yields significant improvements in recognition and retrieval tasks over existing methods, as shown in Sec. \ref{sec:results}.

In summary, our main contributions are:
\begin{itemize}[
  \setlength{\IEEElabelindent}{1pt}%
]
     \item We introduce a new dataset called ObjectsWithStateChange (OWSC)
       that incorporates state changes in object appearance that are
       in addition to the more commonly used transformations related
       to changes in pose and viewpoint. We evaluate several
       state-of-the-art approaches for recognition and retrieval with this dataset. The recognition and retrieval tasks are both at the object-level and at the category-level.
 	\item We propose a new mining strategy that is based on the principles of curriculum learning and that allows our framework to learn discriminative
          object-level representations more effectively that works well even in the presence of state changes for the
          objects.  This strategy leads to performance improvements on
          object-level tasks not only on the proposed ObjectsWithStateChange dataset but also on three other publicly available
          multi-view datasets, ObjectPI, ModelNet-40, and FG3D.
\end{itemize}

\begin{figure}
    \centering
    \includegraphics[width=0.5\textwidth]{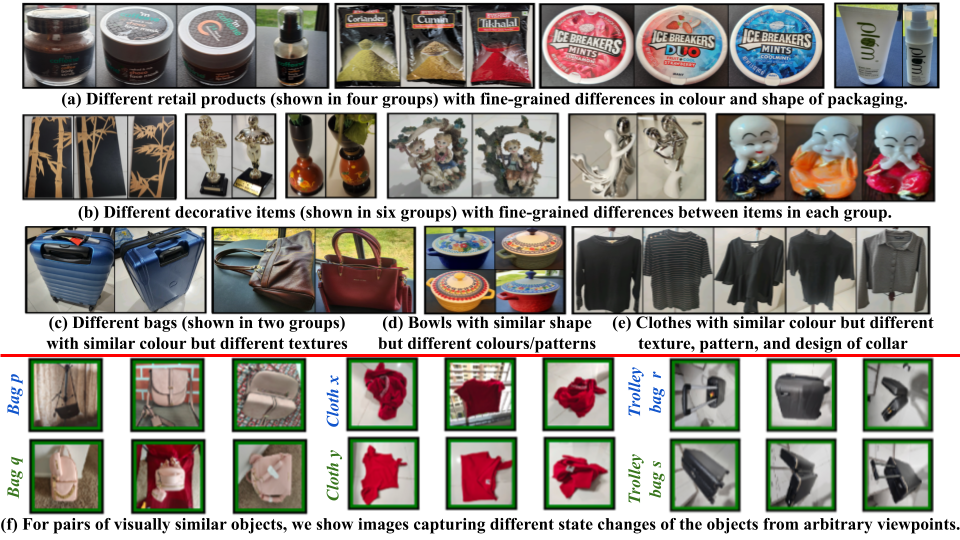}
    \caption{Figs. (a)-(e) show a few examples of visually similar objects from different categories in our dataset for fine-grained recognition and retrieval tasks in the top three rows. Fig. (f) shows samples of pairs of visually similar objects from the same category under various state and pose changes captured from arbitrary views in the bottom two rows. }
    \label{fig:fine-grain_samples}
    \label{fig:fine-grain}
\end{figure}

\begin{figure*}
    \centering
    \begin{minipage}{0.59\textwidth}
    \includegraphics[width=0.97\textwidth]{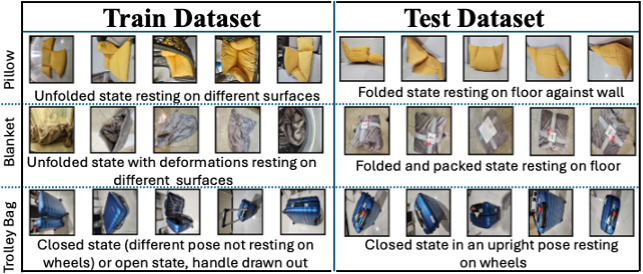}
    \caption{This figure shows samples from the Train and Test splits of our dataset for different objects. The state of the object as well as the background and pose are different in each split for every object. The images are captured from arbitrary viewpoints. }
    \label{fig:splits}
    \end{minipage}
    \hfill
    \vline 
    \hfill 
    \begin{minipage}{0.38\textwidth}
   \includegraphics[width=\textwidth]{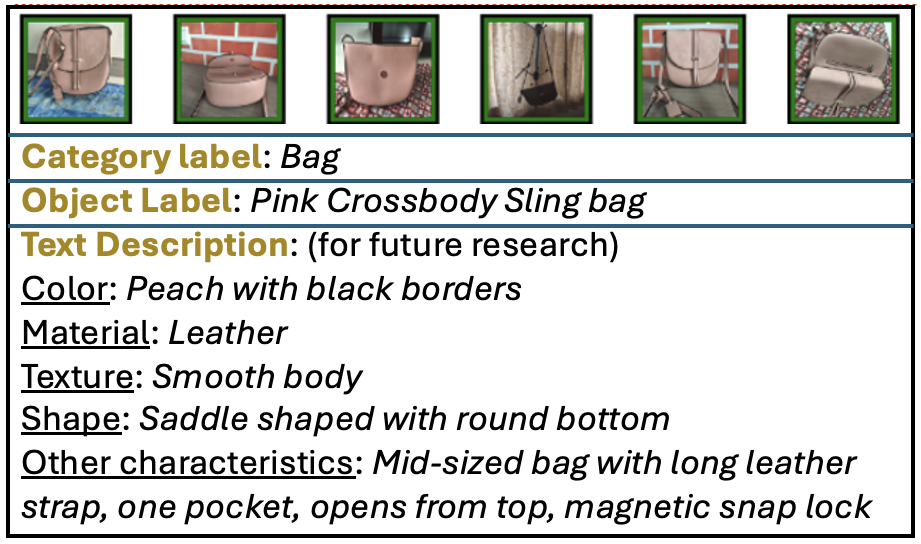}
    \caption{This figure shows the images in various states and poses captured from arbitrary viewpoints, category label, object label, and text description for each object of the OWSC dataset.}
    \label{fig:annotation}
    \end{minipage}
\end{figure*}
\section{Related Work}
{\bf (A) Multi-view Object Datasets: } Prior work has introduced several multi-view object datasets for recognition and retrieval tasks. 
Datasets such as ModelNet-40 \cite{ModelNet40}, and FG3D \cite{FG3D} comprise multi-view images of CAD models of objects. These datasets are intended for learning just the shape of objects and lack color and texture information, that we see in real-world objects. 
Datasets such as ETH-80 \cite{ETH80}, W-RGBD \cite{RGBD}, RPC \cite{RPC2019}, MIRO \cite{RotationNet}, MVP-N \cite{MVP-N} capture real-world objects placed in a controlled environment with uniform illumination and a plain background while the ObjectPI \cite{PIE2019} dataset captures objects in the wild with complex backgrounds. In all these datasets, multi-view data is captured for objects that are placed in particular poses, the data being recorded from a set of pre-defined viewpoints. ObjectNet \cite{ObjectNet} is a test dataset meant for evaluation of category recognition in a real-world setting with arbitrary object backgrounds, rotations, and viewpoints. Toybox \cite{Toybox} is a small video dataset capturing basic transformations such as rotation, translation, and zoom. 

In contrast, our \datasetb dataset captures {\em complex
state changes in addition to alterations in pose, lighting,
background, and viewpoint}, that induce significant
changes in the appearance of the objects within the dataset. Notably,
our dataset includes visually similar objects as shown in Fig. \ref{fig:fine-grain_samples}, making it well-suited
for fine-grained tasks. A comparison of \datasetb with other
datasets is detailed in Table
\ref{tbl:comp_ObjectTransform}. Furthermore, to aid in the learning of
multi-modal representations, we provide text annotations
describing the visual attributes of the objects.

\noindent
{\bf (B) Curriculum Learning: } 
As introduced in \cite{CurriculumICML}, it organizes learning tasks by gradually increasing the complexity of the data samples during the training phase, starting with the easiest cases 
and ending with those that are the most challenging \cite{CurriculumTPAMI}.
As an extension of this basic idea, in the domain of metric learning, Sanakoyeu et al. \cite {DCESMLITPAMI} partition both the data and the embedding space into smaller subsets and learn separate distance metrics within each of the non-overlapping subspaces. Meanwhile, Sarkar et al. \cite{Sarkar_2023_WACV, Sarkar_2022_CVPR} proposed a hierarchical sampling strategy for retrieval tasks, initially sampling negatives from the same high-level category and then selecting harder negatives from the same fine-grained category.

\noindent 
{\bf (C) Pose-invariant Methods: }
Pose-invariant methods \cite{PIE2019} focus on learning category-specific embeddings, representing the object-to-object variations within the same embedding space. However, these approaches struggle to differentiate between objects within the same category, resulting in subpar performance on object-based tasks. To address this, PiRO \cite{PiRO} designed a dual-encoder model to decouple object and category embeddings, achieving improved object recognition and retrieval performance by effectively distinguishing between similar objects within the same category.

In our work, we evaluate the performance of these methods on our new dataset. Additionally, we train the PiRO \cite{PiRO} framework with a curriculum learning strategy. While PiRO randomly samples a pair of objects from the same category, we additionally sample similar objects from the same and other categories. As the training progresses, we sample objects that are harder to distinguish. This approach captures more discriminative object-identity representations to distinguish between visually similar objects, improving performance on object-level tasks (ref. Tables \ref{tbl:curriculum_ablation_OWSC}, \ref{tbl:curriculum_ablation_PICR}).
Also, the dual endoder in PiRO utilizes a single-head self-attention layer for each subspace to aggregate visual features from different views, we adopt a similar architecture with multiple attention layers. This modification allows us to effectively capture invariant visual features under more complex transformations beyond just pose changes (ref. Table \ref{tbl:curriculum_architecture}).

\section{The ObjectsWithStateChange Dataset}
\label{sec:dataset}
\subsection{Dataset Design and Collection}
{\bf Category and Object Selection:} We chose objects for the OWSC dataset such that they satisfied the following two conditions: (a) The objects had significant appearance changes under different state and pose changes; and (b) 
We chose objects that we thought would be particularly challenging because the differences in their images (from different viewpoints for the same category and across categories) appeared to be subtle for fine-grained object recognition and retrieval tasks. Fig. \ref{fig:fine-grain} shows some examples of the images collected for such objects.
The dataset contains images from the following 21 categories: bags, books, bottles, bowls, clothes, cups, decorations, headphones, telephones, pillows, plants, plates, remotes, retail products, toys, ties, towels, trolley bags, tumblers, umbrellas, and vegetables. The number of objects in these categories totals 331. We made sure that all these objects met the two conditions just stated.
For each object, multiple images were collected as described below. 

\noindent
{\bf Dataset Collection: } The OWSC dataset comprises 11328 images of 331 household objects from 21 categories, captured using smartphone cameras under various state changes. Detailed descriptions of these state variations and other transformations specific to each object category, are provided in the supplemental (ref. Table \ref{tbl:state_changes}). Each object in a given state was placed in multiple stable poses on various surfaces, with variations in lighting conditions and backgrounds. Objects were then photographed under these variations from arbitrary viewpoints, capturing a wide range of angles and diverse perspectives. 

During data collection, all images of an object in the {\em same state} captured from multiple viewpoints, including variations in pose, lighting, and background, were grouped together. The dataset was then randomly partitioned into training and test sets with a 3:1 ratio, where one-fourth of the states per object were randomly selected for the test set, and all images of the object in those states were allocated to the test set, while the images from the remaining three-fourths of the states of the same object were assigned to the training set. This ensures that the training and test sets contain images of each object in different states, with no overlap between the two partitions. The training dataset consists of 7,900 images, with approximately 24 images per object, while the test dataset contains 3,428 images, with approximately 10 images per object. This setup allows the model to be evaluated on its ability to learn invariant representations for recognizing and retrieving objects across different states and transformations. Some examples from both splits are shown in Fig. \ref{fig:splits}. The dataset is publicly released at \url{https://github.com/sarkar-rohan/ObjectsWithStateChange}. 

\begin{figure*}
    \centering
    \begin{minipage}{0.33\textwidth}
    \centering
    \includegraphics[width=0.9\textwidth]{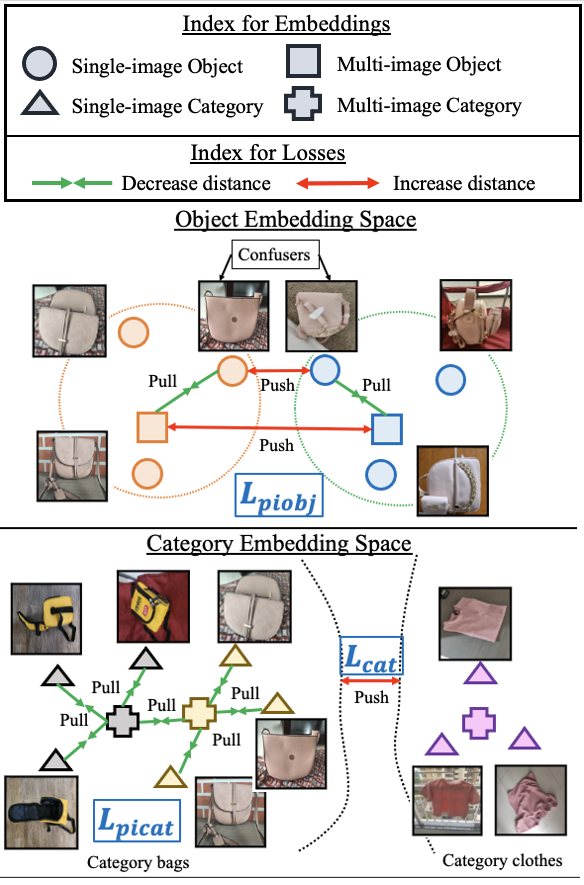}
    \caption{An overview of the ranking losses from \cite{PiRO} described in Sec. \ref{sec:losses} for the object space (top) and the category space (bottom) that we use in our work to learn invariant embeddings.}
    \label{fig:losses}
    \end{minipage}
    \hfill
    \vline 
    \hfill 
    \begin{minipage}{0.65\textwidth}
    \centering
    \includegraphics[width=0.93\textwidth]{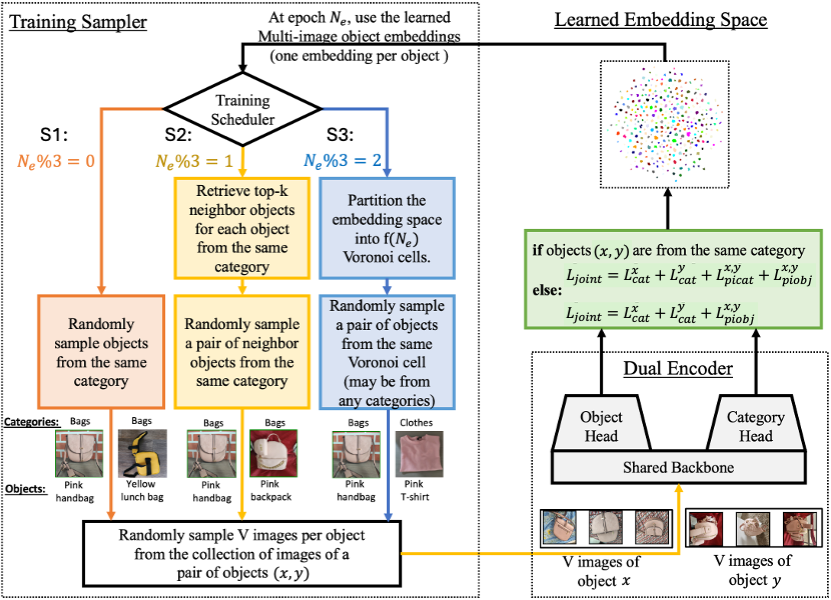}
    \caption{An overview of the proposed training pipeline to learn invariant embeddings. Our curriculum learning strategy uses inter-object distances in the currently learned embedding space to sample pairs of visually similar objects from the same and other categories to train the dual encoder jointly using the losses in Fig. \ref{fig:losses}. As shown in the left, our method switches between three sampling strategies (ref.  Sec. \ref{sec:curriculum}), based on the epoch number $N_e$. }
    \label{fig:framework}
    \end{minipage}
\end{figure*}

\noindent
{\bf Dataset Annotation and Pre-processing:} Each object is annotated with a category label, object label, and a text description describing the visual characteristics of the object to facilitate multi-modal learning, as shown in Fig. \ref{fig:annotation}. The images are captured from arbitrary viewpoints without recording specific pose or viewpoint information. No preprocessing is performed to preserve the natural background of the images, ensuring their suitability for in-the-wild recognition and retrieval.

\subsection{Benchmarking using ObjectsWithStateChange}
\label{sec:benchmark}
\noindent 
{\bf Single-image and Multi-image State-Invariant Recognition and Retrieval Tasks:}
For evaluation using our dataset, we introduce eight state-invariant tasks focused on category- and object-level recognition and retrieval as described below: \\
\textbullet {\em Single-image or multi-image category recognition}: Predict the category label from either a single image or a set of images.\\
\textbullet {\em Single-image or multi-image object recognition}: Predict the object-identity label from either a single image or a set of images.\\ 
\textbullet {\em Single-image or multi-image category retrieval}: Retrieve images of objects belonging to the same category as the query object, using either a single image or a set of images as the query.\\
\textbullet {\em Single-image or multi-image object retrieval}: Retrieve images of the same object-identity as the query object, using either a single image or a set of images as the query.

\noindent 
{\bf Evaluation metrics: }
Classification and retrieval performance are measured in terms of accuracy and mean average precision (mAP).
\section{Proposed Framework for Learning Invariant Embeddings}
\label{sec:proposed_method}
For the work reported in this paper, we adapted the PiRO framework \cite{PiRO} to deal with
the challenges caused by state changes of objects that exhibit similar visual characteristics.
First, Sec. \ref{sec:dual_encoder} provides a brief overview of the network architecture and modifications we make. Next, Sec. \ref{sec:losses} describes the losses in \cite{PiRO} and how they can be used for learning state-invariant embeddings from a collection of object images. Lastly, we elaborate on the motivation and the details of our curriculum learning strategy to mine informative samples for learning discriminative state-invariant embeddings in Sec. \ref{sec:mining}.

\subsection{The Dual-Encoder Architecture}
\label{sec:dual_encoder}

PiRO \cite{PiRO} introduced a {\em dual-encoder model} comprising a shared CNN backbone and two attention heads, each dedicated to learning category and object-identity representations in two distinct embedding spaces. 
In the work described in \cite{PiRO},  using self-attention in their respective embeddings spaces, the heads captured category- and object-specific features that remained invariant to transformations like pose changes.  As we will show later in this paper, the same embeddings also remain invariant to state changes when trained using curriculum learning, as described in Sec. \ref{sec:mining}. 

In our work, we input a set of $V$ unordered images of an object corresponding to its different states and other transformations to the encoder that outputs {\em $V$ single-image embeddings} and {\em an aggregated multi-image embedding} for each embedding space. While PiRO employs a single-headed self-attention layer to aggregate features from different viewpoints, {\em our method uses multiple attention layers}, 
enabling it to effectively capture invariant features under intricate transformations caused by state changes, extending beyond just pose variations.
 This is validated in our ablation study in Table \ref{tbl:curriculum_architecture}.
\subsection{Losses}
\label{sec:losses}

The model is trained using three margin-based losses from \cite{PiRO} to optimize the distances between the images of {\em object pairs} to learn state-invariant object and category embeddings, as follows:

\noindent 
{\bf Object Loss:} As shown in Fig. \ref{fig:losses} (top), $L_{piobj}$ reduces intra-object distances by pulling the embeddings of mutually confusing instances (confusers) between pairs of objects closer to the multi-image embedding of the same object (shown in green arrows), within a margin $\alpha$. Simultaneously, it increases the inter-object distances by separating the confusers and multi-image embeddings of different objects (shown in red arrows) by a margin $\beta$.

\noindent 
{\bf Two Category Losses:} As shown in Fig. \ref{fig:losses} (bottom), 
$L_{picat}$ clusters the embeddings of pairs of objects within the same category (shown in green arrows), ensuring intra-category distances remain within a margin $\theta$. Meanwhile,
$L_{cat}$ enforces inter-category separation by pushing the embeddings from different categories by a margin $\gamma$ (shown in red arrows).

\begin{figure}
    \centering
    \includegraphics[width=0.5\textwidth]{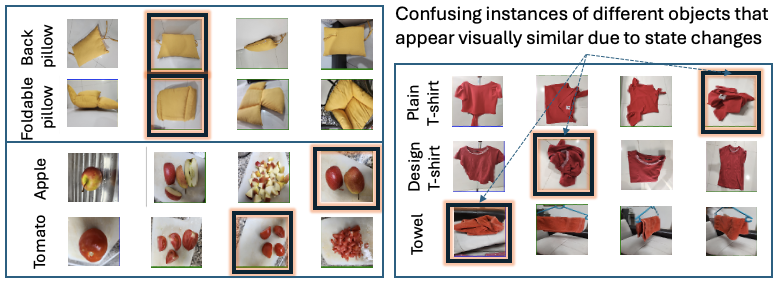}
    \caption{This figure highlights the challenges caused by state changes. Significant variations in the appearance of an object in different states result in high intra-class variance among samples of the same object. More importantly, due to changes in state and pose, instances of distinct objects from the same and different categories can appear similar, making them hard to distinguish, as highlighted by the black bounding boxes. The figure includes examples from our OWSC dataset, such as a back pillow in its original state and another foldable pillow in a folded state, a chopped tomato and apple appearing similar, and so on (left). Notably, two red t-shirts (from the same category) and a red towel (from a different category) are indistinguishable when crumpled (right).}
    \label{fig:state_changes}
\end{figure}

\subsection{Using Curriculum Learning as a Mining Strategy }
\label{sec:mining}
While the overall goal of the learning described in this paper is the same as in \cite{PiRO}, its implementation is made complex by the fact that the objects are now allowed to undergo changes in their state.  That is, we still want to make sure that the {\em embeddings for the same object class are tightly clustered while remaining well separated from those for the other classes}.  However, that goal is difficult to achieve with a direct application of the PiRO framework \cite{PiRO} in the presence of all the additional variability injected by the changes in object state.  Fig. \ref{fig:state_changes} shows some examples of the similarities across classes as created by the changes in state.

 To address these challenges created by the changes in state, we introduce a
new curriculum learning strategy to mine informative samples for training.
  While PiRO randomly samples a pair of objects from the
same category, now we {\em additionally} sample similar objects from
the same {\em and} other categories. This ensures {\em confusing instances of similar objects} (as shown in Fig. \ref{fig:state_changes}) {\em are adequately separated in the embedding space, regardless of their category}. 
Furthermore, as the training progresses, we
sample object pairs that are harder to distinguish to effectively learn
discriminative object-identity representations in order to distinguish
between visually similar objects. For this, we use the inter-object distances in the currently learned embedding space to identify confusing instances of objects (both within and across different categories) and separate them in the embedding space to facilitate learning distinctive features that remain consistent under state changes and other transformations. 

We demonstrate later that our curriculum strategy helps better cluster images of objects undergoing state changes and separate confusing instances in Fig. \ref{fig:UMAP}, thereby leading to performance improvements, as compared to prior methods in Table \ref{tbl:curriculum_ablation_OWSC}. This strategy facilitates learning more 
discriminative embeddings not only for datasets with state changes but
also existing multi-view object datasets, as we will demonstrate later 
in in Table \ref{tbl:curriculum_ablation_PICR}.

\subsubsection{Informative Sample Mining and Training: } 
\label{sec:curriculum}
This section explains our approach to train the dual-encoder model (ref. Sec. \ref{sec:dual_encoder}) using the losses (ref. Sec. \ref{sec:losses}), as illustrated in
Fig. \ref{fig:framework}. These losses optimize the intra-class and inter-class distances between pairs of objects and we provide a detailed explanation of how these object pairs are selected next. 

Training begins by randomly sampling
object pairs from the same category in the first epoch. For each object pair, $V$ images are randomly sampled per object to train the encoder. In subsequent epochs, the learned multi-image object embeddings
 are used to identify similar object pairs from both the same and different
 categories at the start of each epoch to guide the sampling process. As training progresses, our method gradually selects object pairs that are closer in the learned embedding space.

During training, our method switches between three different
sampling strategies to select object pairs, as described below:

\noindent 
{\bf (S1) Randomly sample objects from the same category: }\\
Similar to PiRO \cite{PiRO}, it allows sampling diverse objects from
the same category that may be visually different such as a pink
handbag and a yellow lunch box (both from the category `bags'), helping the encoder learn common
features between them in the category space and discriminative
features in the object space.

\noindent 
{\bf (S2) Randomly sample similar objects from the same category: }\\
While the previous sampling strategy will include similar objects from
the same category, the additional samples generated by this step
increases the proportion of the visually similar pairs of objects
from the same category in order to give greater weightage to such
pairs.  This allows the system to learn fine-grained differences
between the similar-looking objects from the same category such as a pink handbag and a pink backpack from the category
`bags'.  At the
start of the epoch, for each object in the dataset, a list of top-k
neighboring objects in the learned object embedding space is
retrieved using ANN techniques \cite{FAISS}, and, for pair formation, another object is randomly sampled
from this list for training.  

For (S1) and (S2), the dual encoder is jointly trained using the losses mentioned in Section \ref{sec:losses}, as follows 
\begin{equation}
L_{joint} = \frac{1}{|\mathcal{P}_{cat}|}\sum_{(x,y) \in \mathcal{P}_{cat}} L^x_{cat}+L^y_{cat} + L^{x, y}_{picat} + L^{x, y}_{piobj} 
\label{eqn:rand_catg}
\end{equation}
where, $\mathcal{P}_{cat}$ is the set of all object pairs
where each pair $(x, y)$ is sampled from the same category either randomly for (S1)  or based on inter-object distances for (S2).

\noindent 
{\bf (S3) Randomly sample similar objects from any category: }\\
The previous two sampling strategies primarily focused on selecting samples within the same category. However, objects from different categories -- especially those undergoing state changes or transformations -- can also exhibit similar visual characteristics. For example, a red top, towel, cloth bag, and blanket from different categories may appear indistinguishable when crumpled or folded (see Fig.  \ref{fig:state_changes}). This strategy aims to sample such object pairs irrespective of their category. For this, it utilizes the inter-object distances in the learned embedding space to select neighboring objects across all categories, aiming to effectively separate confusing instances of similar objects caused by state changes and other transformations. Furthermore, as training advances, the strategy progressively samples harder-to-distinguish object pairs with lower inter-object distances, regardless of their category. 

Specifically, at the start of the epoch, the object embedding space is partitioned into $f(N_e)$ Voronoi cells using k-means clustering, where $f$ is a function that increases strictly with the current epoch number $N_e$. In our approach, we use a ramp function $f(N_e)=cN_e$, with a constant slope $c$, a lower bound $n_{min}$, and an upper bound $n_{max}$, as detailed further in the next subsection. This leads to progressively dividing the object embedding space into finer partitions as the training advances. For each object in the dataset, another object is randomly sampled from within the same Voronoi cell. As the training progresses, partitions become finer, yielding sampled object pairs that are closer in the object embedding space and share more similar characteristics, making them harder to distinguish.

For (S3), the dual encoder is jointly trained using
\begin{equation}
L_{joint} = \frac{1}{|\mathcal{P}_{part}|}\sum_{(x,y) \in \mathcal{P}_{part}} L^x_{cat}+L^y_{cat} + L^{x, y}_{piobj} 
\label{eqn:rand_part}
\end{equation}
where, $\mathcal{P}_{part}$ is the set of all object pairs
where each pair $(x,y)$ comprises neighboring objects from the same partition in the learned object embedding space at the beginning of the epoch. Note that $L^{x, y}_{picat}$
which clusters the embeddings of a pair of objects from the same category together in the category space, 
is not used as the object pairs $(x,y)$ may be sampled from different categories. 

\subsubsection{Implementation Details: }
For a fair comparison with prior methods \cite{PIE2019,PiRO}, VGG-16 is used as the CNN backbone. The last FC layers are modified to generate 2048-D embeddings and are initialized with random weights. Two-layer single-head self-attention layers are used for each embedding space with a dropout of 0.25. Training involves jointly optimizing the category and object-based losses, as described in Eqs. \ref{eqn:rand_catg}, \ref{eqn:rand_part}. Margins $\alpha=0.25$ and $\beta=1.00$ are set for the object space, while margins $\theta=0.25$ and $\gamma=4.00$ are set for the category space, which are the same as \cite{PiRO}.
 $V$ is set to 12 images per object, and the images are resized to 224$\times$224 and normalized. 
 We use the Adam optimizer with a learning rate of $5e^{-5}$. We train for 150 epochs and use the step scheduler that reduces the learning rate by half after every 30 epochs. 
 For curriculum learning, at the beginning of each epoch an IVFFlat index is built using FAISS \cite{FAISS} using the currently learned multi-image object embeddings (with one aggregated embedding per object). For this, the quantizer used is FlatL2 and the object embedding space is divided into $f(N_e) = max(min(2 \times N_e, n_{max}), n_{min})$ partitions, where $N_e$ is the current training epoch, $n_{min}=8$ and $n_{max}=100$. 

\section{Experimental Results}
\label{sec:results}
\subsection{Evaluation using ObjectsWithStateChange}

\begin{table}[t]
\caption{Comparison of classification and retrieval performance for
  the category and object-based tasks against several state-of-the-art
  pose-invariant methods using the new ObjectsWithStateChange dataset. SV
  and MV denote either a single image or multiple images used at
  the time of inference respectively. The results indicate that our
  method outperforms prior methods on all tasks. }
\centering
\setlength{\tabcolsep}{2pt}
\scriptsize
\begin{tabular}{@{}lccccccccccc@{}}
\toprule
\multicolumn{1}{l}{\multirow{3}{*}{Method}} & \multirow{3}{*}{\begin{tabular}[c]{@{}c@{}}Embed.\\ Space\end{tabular}} & \multicolumn{5}{c}{Classification (Accuracy \%)} & \multicolumn{5}{c}{Retrieval (mAP \%)} \\ \cmidrule(l){3-12} 
\multicolumn{1}{c}{} &  & \multicolumn{2}{c}{Category} & \multicolumn{2}{c}{Object} & \multirow{2}{*}{Avg.} & \multicolumn{2}{c}{Category} & \multicolumn{2}{c}{Object} & \multirow{2}{*}{Avg.} \\ \cmidrule(lr){3-6} \cmidrule(lr){8-11}
\multicolumn{1}{c}{} &  & SV & MV & SV & MV &  & SV & MV & SV & MV &  \\ \midrule
PI-CNN \cite{PIE2019} & Single & 82.57 & 84.29 & 25.37 & 26.74 & 54.74 & 84.75 & 86.87 & 11.35 & 36.89 & 54.96 \\
PI-Proxy \cite{PIE2019} & Single & 84.51 & 86.86 & 27.10 & 28.85 & 56.83 & 86.69 & 89.33 & 12.27 & 39.83 & 57.03 \\
PI-TC \cite{PIE2019} & Single & 81.55 & 84.29 & 42.04 & 52.12 & 65.00 & 84.21 & 88.52 & 22.02 & 61.72 & 64.12 \\
PiRO \cite{PiRO} & Dual & 88.07 & 92.30 & 69.01 & 74.78 & 81.04 & 89.44 & 94.16 & 68.97 & 81.86 & 83.61 \\ \midrule
{\bf Ours} & Dual & \textbf{89.39} & \textbf{93.36} & \textbf{77.57} & \textbf{87.77} & \textbf{87.02} & \textbf{90.70} & \textbf{94.56} & \textbf{79.22} & \textbf{91.17} & \textbf{88.91} \\ \bottomrule
\end{tabular}
\label{tbl:result_overall}
\end{table}
We compare our method with various state-of-the-art pose-invariant methods using our OWSC dataset. 
The prior methods were originally designed to learn pose-invariant embeddings from multi-view images, but can be adapted to learn invariant embeddings from object image collections with diverse transformations from arbitrary views as they cluster multiple images of the same object together without needing explicit pose, viewpoint, or state information during training. 

\noindent
{\em Single Embedding Space Methods: } Ho et al. \cite{PIE2019} proposed three pose-invariant methods PI-CNN, PI-Proxy, and PI-TC that focused primarily on learning category-specific embeddings, where the object-to-object variations are represented by variations in embedding vectors within the same embedding space. Specifically, these methods clustered single-view embeddings close to the multi-view embedding of the same object, that is clustered close to a learned proxy embedding of its category.
However, the embeddings of different objects within the same category are not explicitly separated, yielding poor performance on object-based tasks. 
\begin{figure}[t]
    \centering
    \includegraphics[width=0.5\textwidth]{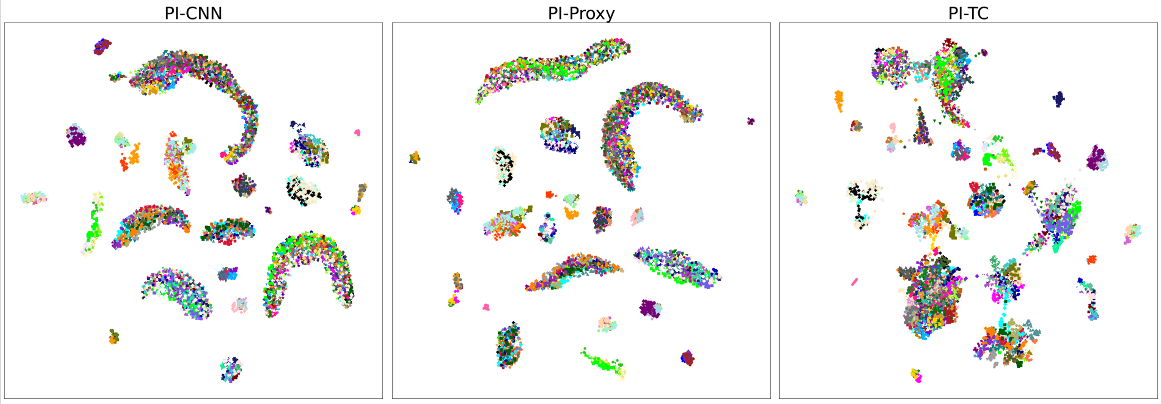}
    \includegraphics[width=0.5\textwidth]{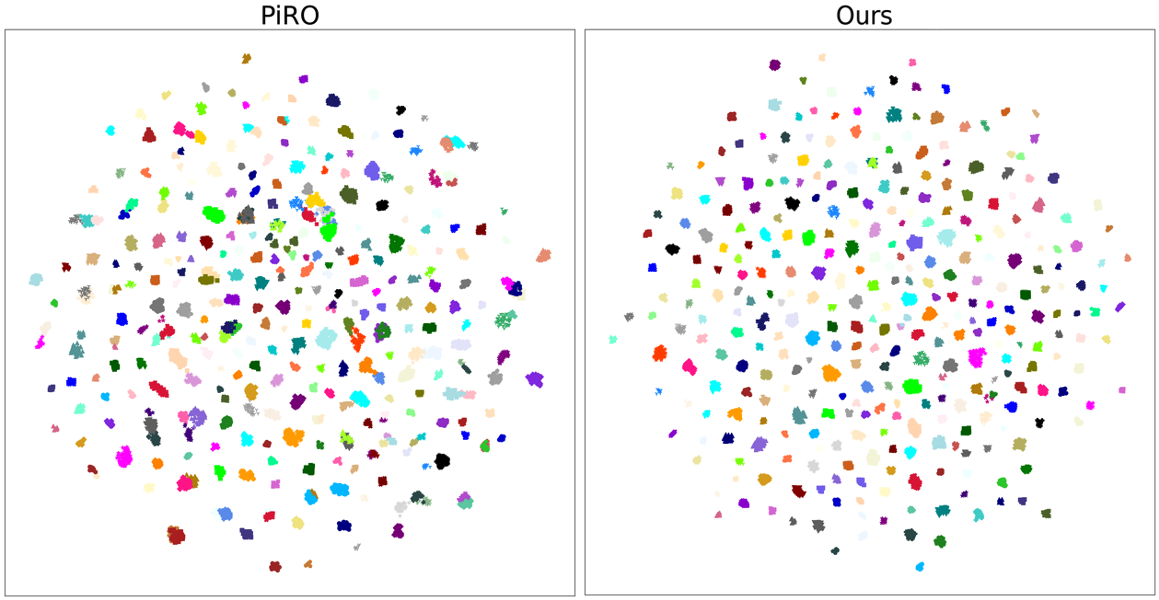}
    \caption{Comparison of the UMAP visualization of the embeddings of the 331 objects in our \datasetb dataset learned using different methods. The embeddings for each object are shown using a unique color and pattern. The figures in the top row show the embeddings learned by the pose-invariant methods \cite{PIE2019} that learn a single embedding (SE) space while the figures in the bottom row show the embeddings for methods learning a dual embedding (DE) space. It can be seen that the SE methods cannot separate objects from the same category. In comparison, the dual embedding (DE) methods can better separate the different objects despite the state changes and transformations. When comparing PiRO (bottom-left) and our proposed method  (bottom-right), we see that the embeddings for different objects are better separated using our method. }
    \label{fig:UMAP}
\end{figure}

\noindent
{\em Dual Embedding Space Methods:}
To address this, PiRO \cite{PiRO} designed a dual-encoder model that decouples category and object representations in separate embedding spaces using margin-based losses specifically designed for each embedding space. Our proposed framework also learns dual embeddings by extending PiRO \cite{PiRO} using a curriculum learning approach, as described in Sec. \ref{sec:proposed_method}.

We train these methods using our OWSC dataset and compare the results in Table \ref{tbl:result_overall}. We observe that our method outperforms all the prior methods. This is because our method can separate fine-grained objects under different state changes and other transformations more effectively than prior methods as demonstrated using UMAP visualization of the embeddings in Fig. \ref{fig:UMAP}.  Despite the high intra-class variance and similarity across different object classes caused by state changes, we can observe that the embeddings of each object are better clustered and separated from other objects for our method. This is primarily due to the curriculum learning strategy that not only separates objects from the same category but also separates visually similar objects from other categories. Also, in the next subsection, we observe that using multiple attention layers further improves performance as it helps capture invariant features under complex transformations and state changes. We provide detailed ablation of both the curriculum learning and the architectural changes in Section \ref{sec:ablations}. We further observe that our curriculum learning approach improves performance on object-level tasks for three other multi-view datasets (ObjectPI, ModelNet, and FG3D) in Table \ref{tbl:curriculum_ablation_PICR}.

\begin{table}[t]
\caption{Ablation study to understand the effect of sampling training examples. PiRO \cite{PiRO} randomly sampled a pair of objects from the same category, whereas our method additionally samples a pair of neighboring objects from the same and other categories. We observe that our curriculum learning strategy improves recognition and retrieval performance on object-level tasks as compared to the state-of-the-art \cite{PiRO}. }
\centering
\scriptsize
\setlength{\tabcolsep}{2pt}
\begin{tabular}{@{}lccccccccccc@{}}
\toprule
\multirow{3}{*}{Dataset} & \multicolumn{1}{c}{\multirow{3}{*}{\begin{tabular}[c]{@{}c@{}}Sampling \\ Method\end{tabular}}} & \multicolumn{5}{c}{Classification (Accuracy \%)} & \multicolumn{5}{c}{Retrieval (mAP \%)} \\ \cmidrule(l){3-12} 
 & \multicolumn{1}{c}{} & \multicolumn{2}{c}{Category} & \multicolumn{2}{c}{Object} & \multirow{2}{*}{Avg.} & \multicolumn{2}{c}{Category} & \multicolumn{2}{c}{Object} & \multirow{2}{*}{Avg.} \\ \cmidrule(lr){3-6} \cmidrule(lr){8-11}
 & \multicolumn{1}{c}{} & SV & MV & SV & MV &  & SV & MV & SV & MV &  \\ \midrule
\multirow{2}{*}{\begin{tabular}[c]{@{}l@{}}OWSC \end{tabular}} & Same category & \textbf{87.1} & \textbf{91.2} & 68.7 & 76.4 & 80.9 & 88.5 & \textbf{93.2} & 68.8 & 82.9 & 83.4 \\ \cmidrule(l){2-12} 
 & Curriculum (Ours) & 86.9 & 89.4 & \textbf{76.6} & \textbf{83.4} & \textbf{84.1} & \textbf{88.6} & 92.4 & \textbf{78.0} & \textbf{88.0} & \textbf{86.8} \\ \midrule
\end{tabular}
\subcaption{Results on single and multi-image state-invariant recognition and retrieval tasks on our ObjectsWithStateChange (OWSC) dataset.}
\label{tbl:curriculum_ablation_OWSC}
\setlength{\tabcolsep}{2pt}
\begin{tabular}{@{}lccccccccccc@{}}
\toprule
\multirow{3}{*}{Dataset} & \multicolumn{1}{c}{\multirow{3}{*}{\begin{tabular}[c]{@{}c@{}}Sampling \\ Method\end{tabular}}} & \multicolumn{5}{c}{Classification (Accuracy \%)} & \multicolumn{5}{c}{Retrieval (mAP \%)} \\ \cmidrule(l){3-12} 
 & \multicolumn{1}{c}{} & \multicolumn{2}{c}{Category} & \multicolumn{2}{c}{Object} & \multirow{2}{*}{Avg.} & \multicolumn{2}{c}{Category} & \multicolumn{2}{c}{Object} & \multirow{2}{*}{Avg.} \\ \cmidrule(lr){3-6} \cmidrule(lr){8-11}
 & \multicolumn{1}{c}{} & SV & MV & SV & MV &  & SV & MV & SV & MV &  \\ \midrule
\multirow{2}{*}{\begin{tabular}[c]{@{}l@{}}Object\\ PI\end{tabular}} & Same category & \textbf{71.3} & 83.7 & 92.7 & 98.0 & 86.4 & \textbf{65.7} & \textbf{83.4} & 81.0 & 99.0 & 82.3 \\ \cmidrule(l){2-12} 
 & Curriculum (Ours) & 70.3 & \textbf{85.7} & \textbf{96.2} & \textbf{99.0} & \textbf{87.8} & 64.9 & 82.4 & \textbf{86.3} & \textbf{99.3} & \textbf{83.2} \\ \midrule
\multirow{2}{*}{\begin{tabular}[c]{@{}l@{}}Model\\Net40\end{tabular}} & Same category & \textbf{85.2} & 88.9 & 94.0 & 96.9 & 91.2 & \textbf{79.7} & \textbf{86.1} & 84.0 & 98.2 & 87.0 \\ \cmidrule(l){2-12} 
 & Curriculum (Ours) & 84.7 & \textbf{89.0} & \textbf{95.5} & \textbf{97.5} & \textbf{91.7} & 79.6 & 85.8 & \textbf{87.9} & \textbf{98.6} & \textbf{88.0} \\ \midrule
 \multirow{2}{*}{\begin{tabular}[c]{@{}l@{}}FG3D\end{tabular}} & Same category & \textbf{79.0} & \textbf{81.8} & 83.1 & 91.5 & 83.9 & \textbf{68.1} & \textbf{74.4} & 73.0 & 95.5 & 77.8 \\ \cmidrule(l){2-12} 
 & Curriculum (Ours) & 78.3 & 81.3 & \textbf{84.7} & \textbf{92.7} & \textbf{84.3} & 67.6 & 73.7 & \textbf{77.3} & \textbf{96.1} & \textbf{78.7} \\ \midrule
\end{tabular}
\subcaption{Results on single and multi-view pose-invariant recognition and retrieval tasks on three publicly available multi-view datasets. }
\label{tbl:curriculum_ablation_PICR}
\end{table}
\subsection{Ablation Studies}
\label{sec:ablations}
\noindent 
{\bf (A) Curriculum Learning: } PiRO uses multi-view images of
a pair of objects randomly sampled from the same category, whereas our new
method additionally samples images of a pair of neighboring objects
from the same category and other categories. The question now is whether the additional computational burden entailed by this sampling (ref. Appendix \ref{sec:time_complexity}) is worth the effort. This ablation is an answer to that question.

We compare the
performance of the two sampling strategies using our
ObjectsWithStateChange dataset on transformation-invariant recognition
and retrieval tasks in Table \ref{tbl:curriculum_ablation_OWSC}. We
additionally also compare performance on pose-invariant recognition
and retrieval tasks using three publicly available multi-view datasets
(ObjectPI \cite{PIE2019}, ModelNet-40 \cite{ModelNet40}, and FG3D \cite{FG3D}) in Table \ref{tbl:curriculum_ablation_PICR}. For a fair
comparison, the same architecture in \cite{PiRO} with a single-layer and single-head
attention layer is used in both cases.

We observe that our strategy improves recognition and retrieval performance on object-based tasks. Specifically, in Table \ref{tbl:curriculum_ablation_OWSC}, it improves single-view object recognition by 7.9\% and single-view object 
retrieval by 9.2\% over sampling objects randomly from the same category \cite{PiRO}. Similarly, in Table \ref{tbl:curriculum_ablation_PICR}, we observe that our sampling strategy improves single-view object recognition accuracy by 3.5\% on ObjectPI, 1.5\% on ModelNet, and 1.6\% on FG3D, and single-view object retrieval mAP by 5.3\% on ObjectPI, 3.9\% on ModelNet, and 4.3\% on FG3D. 
For multi-view object recognition and retrieval tasks, we see a significant improvement on our OWSC dataset in Table \ref{tbl:curriculum_ablation_OWSC} and marginal improvements on ObjectPI, ModelNet-40, and FG3D datasets in Table \ref{tbl:curriculum_ablation_PICR}. 

However, we observe slight performance degradation on some of the category-based tasks. We conjecture that our curriculum learning approach focuses more on separating visually similar objects to improve performance on object-level tasks than category-level tasks. While these samples are harder to separate in the object embedding space, they are much easier to pull together in the category embedding space as they have very similar attributes,
and are hence not informative for learning category-specific features.

\begin{table}[t]
\caption{Ablation study to understand the effect of the number of heads and self-attention layers in the architecture. Using two layers and a single head yields the best performance on all tasks. }
\centering
\scriptsize
\setlength{\tabcolsep}{4pt}
\begin{tabular}{@{}cccccccccccc@{}}
\toprule
\multicolumn{2}{c}{\multirow{2}{*}{Encoder}} & \multicolumn{5}{c}{Classification (Accuracy \%)} & \multicolumn{5}{c}{Retrieval (mAP \%)} \\ \cmidrule(l){3-12} 
\multicolumn{2}{c}{} & \multicolumn{2}{c}{Category} & \multicolumn{2}{c}{Object} & \multirow{2}{*}{Avg.} & \multicolumn{2}{c}{Category} & \multicolumn{2}{c}{Object} & \multirow{2}{*}{Avg.} \\ \cmidrule(r){1-6} \cmidrule(lr){8-11}
nHead & nLayer & SV & MV & SV & MV &  & SV & MV & SV & MV &  \\ \midrule
1 & 1 & 86.9 & 89.4 & 76.6 & 83.4 & 84.1 & 88.6 & 92.4 & 78.0 & 88.0 & 86.8 \\
2 & 1 & 87.8 & 91.2 & 75.2 & 84.3 & 84.6 & 89.3 & 93.7 & 76.8 & 88.9 & 87.2 \\
4 & 1 & 89.2 & 92.2 & 75.4 & 85.5 & 85.6 & 90.4 & 94.4 & 77.3 & 89.8 & 88.0 \\
8 & 1 & 88.7 & 91.8 & 76.6 & 85.2 & 85.6 & 90.0 & 93.7 & 78.1 & 89.2 & 87.8 \\
1 & 2 & 89.2 & \textbf{93.7} & 77.0 & \textbf{86.7} & \textbf{86.6} & 90.5 & 94.7 & 78.7 & 90.2 & 88.5 \\
2 & 2 & 89.4 & 92.2 & 77.5 & 85.8 & 86.2 & \textbf{90.7} & 94.2 & 79.1 & 89.6 & 88.4 \\
4 & 2 & 88.9 & 93.1 & 77.3 & 85.2 & 86.1 & 90.6 & {\bf 95.0} & 78.9 & 89.1 & 88.4 \\
8 & 2 & \textbf{89.7} & 91.8 & \textbf{78.6} & 86.4 & {\bf 86.6} & 90.6 & 93.3 & \textbf{80.1} & \textbf{90.4} & \textbf{88.6} \\ \bottomrule
\end{tabular}
\label{tbl:curriculum_architecture}
\end{table}
\begin{figure*}[t]
    \centering
    \begin{minipage}[t]{0.22\textwidth}
    \includegraphics[width=\textwidth]{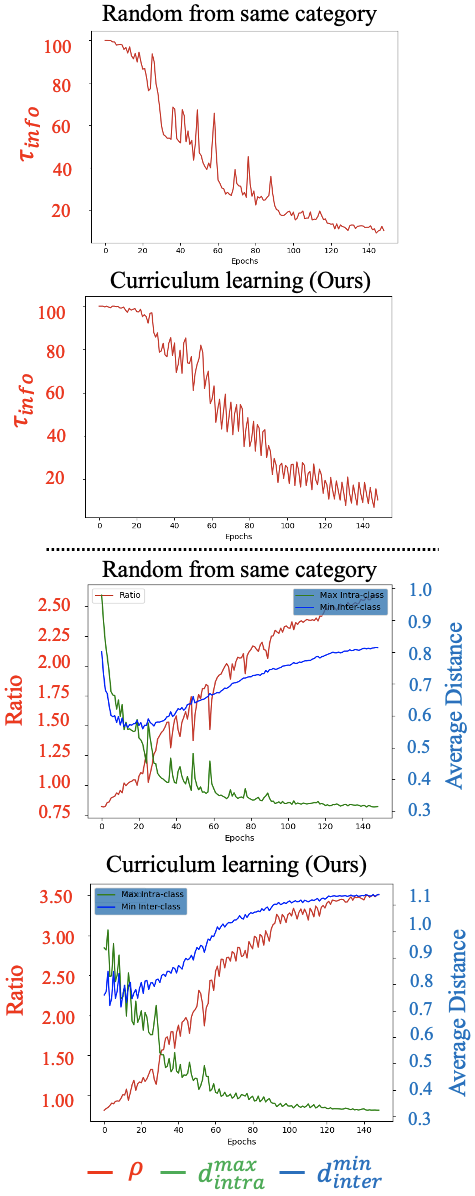}
    \caption{Variation of $\tau_{info}$ (top), and ratio $\rho$, and distances $d^{max}_{intra}$ and $d^{min}_{inter}$ (bottom) vs epochs during training on our \datasetb dataset.}
    \label{fig:distances}
    \end{minipage}
    \hfill
    \vline
    \hfill
    \begin{minipage}[t]{0.77\textwidth}
    \includegraphics[width=\textwidth]{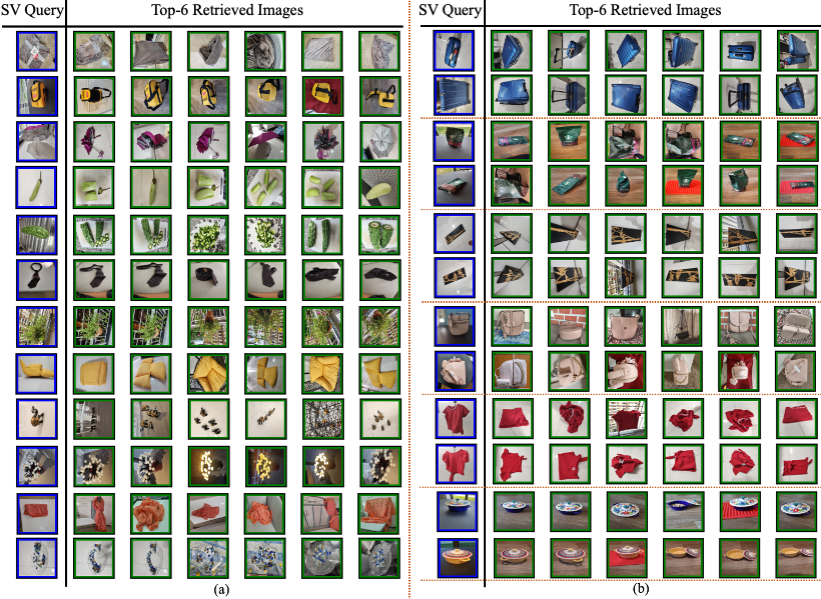}
    \caption{This figure (best viewed zoomed in) presents the results of single-image object retrieval from our dataset using our proposed method. The blue bounding box highlights the query image, while the green bounding box indicates the successful retrieval of an image of the correct object-identity. Our method demonstrates accurate retrieval of images of the same object under different transformations in (a) on the left side and effectively distinguishes the correct object from similar-looking ones as shown in consecutive rows in (b) on the right side. }
    \label{fig:Retrieval}
    \end{minipage}
\end{figure*}
\noindent
{\bf (B) Architecture: } As compared to \cite{PiRO} that used a single attention layer with a single head, we experimented with using multiple attention layers and multi-head attention to capture invariant visual features under complex transformations beyond just pose changes in the case of multi-view datasets. The results in Table \ref{tbl:curriculum_architecture} indicate that using a single attention layer and multiple heads improves multi-image recognition and retrieval performance marginally. Using two attention layers with a single head yields the best performance. However, further increasing the number of heads does not lead to any significant improvements for category-level tasks but yields marginal improvements on object-level tasks.

\noindent
{\bf (C) Optimization of Intra-Object and Inter-Object Distances During Training: }
In Fig. \ref{fig:distances} (top), we plot the percentage of training samples that are informative ($\tau_{\text{\it info}}$ in red). A training sample is considered informative if the loss $L_{piobj}$ for that sample is non-zero. Being a margin-based loss, if the training samples of different objects are already well separated or if the training samples of the same object are already well clustered then the loss $L_{piobj}$ for that sample will be zero and the sample is uninformative.  If we compare $\tau_{\text{\it info}}$ at any given epoch, the curriculum learning approach generates more informative samples than sampling objects from the same category, especially during the initial stages of training. 

Similar to \cite{PiRO}, we plot the maximum intra-class distance ($d^{max}_{intra}$ in green), the minimum inter-class distance between object-identity classes ($d^{min}_{inter}$ in blue), and the ratio $\rho=\frac{d^{min}_{inter}}{d^{max}_{intra}}$ (in red) during training in Fig. \ref{fig:distances} (bottom). These metrics are used to monitor the compactness and separability of object-identity classes. Specifically, a lower $d^{max}_{intra}$, and higher $d^{min}_{inter}$ and $\rho$ values would indicate embeddings of the same object-identity class are well clustered and separated from embeddings of other object-identity classes.
We observe that both sampling methods decrease $d^{max}_{intra}$ to the same extent. However, the curriculum learning strategy increases $d^{min}_{inter}$ and thereby increases $\rho$ to a greater extent than random sampling.  Therefore, the object-identity classes are better separated (indicated by the blue lines in the bottom figure), and this results in improved performance on object-level tasks.

\noindent
{\bf (D) Retrieval Results: } The single-image object retrieval results on our dataset presented in Fig. \ref{fig:Retrieval}, show that our method can accurately retrieve images of the same object under various transformations in (a) on the left side and also effectively distinguish the correct object from similar ones in (b) on the right side. Figs. \ref{fig:SV_RETR1}, \ref{fig:SV_RETR2} in the supplemental show more single-image retrieval results of visually similar objects from each category in OWSC.

\section{Conclusion}
We present ObjectsWithStateChange, a novel dataset that, in addition
to capturing the more commonly used transformations like those related
to pose and viewpoint changes, also includes changes in the state of
the objects.  We evaluate several state-of-the-art methods across
eight category-level and object-level classification and retrieval tasks on
this dataset. Additionally, we introduce an example mining strategy to
learn more discriminative object-level representations for
fine-grained tasks. This approach not only boosts performance on our
dataset but also on three other multi-view datasets available
publicly.\\
\noindent
{\bf Broader Impact and Future Work:} Firstly, we introduce a novel dataset that captures
complex state changes of objects that can be used for category-level
and fine-grained object-level recognition and retrieval. This could
benefit several real-world applications such as automatic checkout
systems \cite{CheckSoft, peopcentsys, RPC2019} 
and robotic systems
that deal with object state changes upon interaction. Secondly, it can serve as a real-world test dataset for assessing the robustness of representations learned to state and pose changes, as we show in the supplement (ref. Appendix \ref{sec:cross-dataset}). Lastly, we provide text annotations that can be used to learn
invariant multi-modal representations \cite{CLIP, 4M, MV-CLIP} 
and can be used for generalized zero-shot evaluation and visual question answering, which we aim to pursue in future work.

\appendices
\renewcommand{\thetable}{\Alph{section}\arabic{table}}

\section{State changes of different objects during data collection}
A comprehensive list of all possible state changes and other transformations for objects from each category is provided in Table \ref{tbl:state_changes}. Depending on the physical characteristics of the specific object, all the transformations may not be possible for each object, and images are collected if such state changes are possible for the objects of interest. 

\begin{table*}[th]
\caption{ This table mentions all possible state changes of the objects from 21 categories that are considered during the data collection. }
\centering
\scriptsize
\begin{tabular}{@{}ll@{}}
\toprule
\textbf{Category} & \multicolumn{1}{c}{\textbf{State changes and other transformations}} \\ \midrule
\textit{\textbf{bags}} & \begin{tabular}[c]{@{}l@{}}place flat on a surface, against wall, hang it on a hook, open and close bag, empty bag and fill it to change shape\end{tabular} \\ \midrule
\textit{\textbf{books}} & place flat on surfaces, open and close books and place with cover up \\ \midrule
\textit{\textbf{bottles}} & open and close lid, fill liquid \\ \midrule
\textit{\textbf{bowls}} & \begin{tabular}[c]{@{}l@{}}place normally and up-side down, place objects/food inside, open and close lid (if bowl has lid)\end{tabular} \\ \midrule
\textit{\textbf{clothes}} & put on a hanger, place on floor in a heap, fold properly, hang outside for drying \\ \midrule
\textit{\textbf{cups}} & place normally and up-side down, place liquid (if possible) \\ \midrule
\textit{\textbf{decorations}} & \begin{tabular}[c]{@{}l@{}}place in different stable poses, capture from viewpoints with significant change in appearance, place in complicated backgrounds, and different lighting conditions \\ (especially ones that have reflective surfaces/have light fixtures), change the arrangement of detachable items (if possible)\end{tabular} \\ \midrule
\textit{\textbf{headphones}} & \begin{tabular}[c]{@{}l@{}}change orientations of microphone, change how the wire is placed, lie flat on table and in an upright position\end{tabular} \\ \midrule
\textit{\textbf{telephones}} & change receiver position and orientation, place hanging on wall and flat on table \\ \midrule
\textit{\textbf{pillows}} & \begin{tabular}[c]{@{}l@{}}place on floor and against wall, put force on pillow to deform in several ways, fold (if possible)\end{tabular} \\ \midrule
\textit{\textbf{plants}} & \begin{tabular}[c]{@{}l@{}}move (only if light and not fixed), different lighting conditions (sunlight/cloudy/night), with and without flowers (when applicable)\end{tabular} \\ \midrule
\textit{\textbf{plates}} & place normally and up-side down, place objects/food on surface \\ \midrule
\textit{\textbf{remotes}} & place in different orientations, place in different backgrounds \\ \midrule
\textit{\textbf{retail products}} & \begin{tabular}[c]{@{}l@{}}place in different poses and backgrounds, change the appearance by removing detachable components, opening cap/lid and placing it beside the object (if possible),\\ deform objects (if possible)\end{tabular} \\ 
\midrule 
\textit{\textbf{toys}} & \begin{tabular}[c]{@{}l@{}}place in different poses and backgrounds, change the appearance by removing detachable components (if possible), deform objects (if possible)\end{tabular} \\ \midrule
\textit{\textbf{ties}} & \begin{tabular}[c]{@{}l@{}}tie a knot and place normally and in an arbitrary pose, open knot and place normally and in arbitrary pose, roll the tie\end{tabular} \\ \midrule
\textit{\textbf{towels}} & \begin{tabular}[c]{@{}l@{}}fold and place on a surface, fold and place on a hanger, put on a surface in a heap (with different types of deformations), place on arbitrary-shaped surfaces\end{tabular} \\ \midrule
\textit{\textbf{trolley bags}} & \begin{tabular}[c]{@{}l@{}}place on different stable poses (upright on wheels/flat on ground, on different surfaces along length/breadth),  with the handle inside/drawn out,\\ open the bag and place it on the floor and in an upright condition\end{tabular} \\ \midrule
\textit{\textbf{tumblers}} & \begin{tabular}[c]{@{}l@{}}place normally and upside down, fill liquid (if possible), put complicated backgrounds (especially for transparent ones), place in different lighting conditions \\(especially for those with reflective surfaces)\end{tabular} \\ \midrule
\textit{\textbf{umbrellas}} & \begin{tabular}[c]{@{}l@{}}open state, closed state with/without tying, fold (if possible), handle inside/drawn out, different backgrounds, folded and hanging\end{tabular} \\ \midrule
\textit{\textbf{vegetables}} & \begin{tabular}[c]{@{}l@{}}raw state, arrange multiple items in different ways, place on different utensils, peel skin (if possible), chop into large chunks (if possible), chop finely (if possible)\end{tabular} \\ \bottomrule
\end{tabular}
\label{tbl:state_changes}
\end{table*}

\section{Time Complexity Analysis}
\label{sec:time_complexity}
For this analysis, we assume there are \( N^O \) objects distributed across \( N^C \) categories in the dataset. The distribution of objects per category is assumed to be uniform, with each category containing an average of \( N^O_{C} = \frac{N^O}{N^C} \) objects. Our goal is to examine how the proposed sampling strategy scales with the number of categories and objects. It is important to note that to reduce the complexity of determining neighboring objects in the embedding space, we use a single aggregated embedding for each object extracted from our encoder, regardless of the number of images associated with it. So, the number of images per object is not included in this analysis.

\subsection{Analysis of the Proposed Sampling Strategy}

At the beginning of each epoch, our algorithm utilizes either of the three distinct sampling strategies:

\noindent {\bf (S1)} For each object, we randomly select another object from the same category. This operation has a time complexity of $\mathcal{O}(1)$. When applied to the entire dataset, the overall time complexity is $\mathcal{O}(N^O)$.

\noindent {\bf (S2)} For each object, we construct a list of similar objects within the same category and randomly sample an object pair from this list. This requires finding approximate all-nearest-neighbors \cite{AllNN1989} for all objects in the same category, which incurs a complexity of $\mathcal{O}(N^O_C \log N^O_C)$ for each category using efficient ANN techniques \cite{FAISS}. Consequently, the total time complexity for the entire dataset is $\mathcal{O}(N^C N^O_C \log N^O_C) = \mathcal{O}(N^O \log N^O_C)$.

\noindent {\bf (S3)} 
This strategy involves partitioning the embedding space using k-means clustering and building an inverted index, which has a time complexity of $\mathcal{O}(KN^O I)$, where $K$ representing the number of clusters and $I$ denoting the number of iterations are constants. For each object, we randomly sample another object from the same partition, which has a time complexity of $\mathcal{O}(1)$. Thus, for the entire dataset, the overall time complexity for partitioning and sampling object pairs is $\mathcal{O}(KN^O I) \approx \mathcal{O}(N^O)$, since both $K$ and $I$ are constants independent of $N^O$ or $N^C$.

Overall, the average time complexity of the algorithm is dominated by the second sampling strategy, yielding $\mathcal{O}(N^O \log N^O_C)$ for the entire dataset comprising $N^O$ objects. {\em For each object, the time complexity of finding another object for comparison is therefore $\mathcal{O}(\log N^O_C)$, which is sub-linear with respect to the number of objects per category in the dataset.}

\subsection{Comparison with Other Methods:}
In this section, we compare the time complexity {\em per object} for different methods. 
The state-of-the-art method, PiRO \cite{PiRO}, randomly selects another object from the same category for each object, resulting in a time complexity of $\mathcal{O}(1)$ per object. 
In contrast, prior work (PI-TC) \cite{PIE2019} compares each object with the {\em nearest} multi-view embedding of the object and the {\em nearest} proxy embedding for the categories, which incurs a complexity of $\mathcal{O}(N^C + N^O) = \mathcal{O}(N^O)$ per object, given that $N^C < N^O$.

Our method employs different sampling strategies across various epochs. Sampling strategies S1 and S3 share the same complexity as PiRO \cite{PiRO}, assuming that $K$ and $I$ remain small relative to $N^O$. However, S2 exhibits a higher complexity than PiRO due to the additional $log N^O_C$ term. For most datasets, $N^O_C$ is relatively small, so the logarithmic term does not significantly impact overall complexity. Nevertheless, for datasets with a large number of objects per category, sampling strategy S2 would have higher complexity than PiRO, although it will still have better complexity than PI-TC \cite{PIE2019}.

\section{Cross-Dataset Evaluation for Assessing Robustness to State Changes and Other Transformations}
\label{sec:cross-dataset}

\begin{table}[t]
\caption{Results of cross-dataset performance evaluation. Model is trained on ObjectPI and tested on the ObjectsWithStateChange dataset with 5 seen and 16 unseen categories and all unseen objects. Even for the seen categories, the object transformations and state changes are unseen, and the viewpoints are arbitrary. }
\centering
\scriptsize
\setlength{\tabcolsep}{3pt}
\begin{tabular}{@{}ccccccccccc@{}}
\toprule
\multicolumn{1}{c}{\multirow{3}{*}{\begin{tabular}[c]{@{}c@{}}Sampling \\ Method\end{tabular}}} & \multicolumn{5}{c}{Classification (Accuracy \%)} & \multicolumn{5}{c}{Retrieval (mAP \%)} \\ \cmidrule(l){4-11} 
& \multicolumn{2}{c}{Category} & \multicolumn{2}{c}{Object} & \multirow{2}{*}{Avg.} & \multicolumn{2}{c}{Category} & \multicolumn{2}{c}{Object} & \multirow{2}{*}{Avg.} \\ \cmidrule(lr){2-5} \cmidrule(lr){7-10}
 \multicolumn{1}{c}{} & SV & MV & SV & MV &  & SV & MV & SV & MV &  \\ \midrule
 Same category & \textbf{65.3} & \textbf{64.7} & 43.7 & 45.9 & 54.9 & \textbf{30.2} & \textbf{42.7} & 17.9 & 55.1 & 36.5 \\ \cmidrule(l){1-11} 
Curriculum (Ours) & 64.2 & 63.4 & \textbf{46.5} & \textbf{49.6} & \textbf{55.9} & 29.6 & 42.2 & \textbf{20.8} & \textbf{58.6} & \textbf{37.8} \\ \midrule
\end{tabular}
\label{tbl:zero-shot}
\end{table}

Cross-dataset evaluation involves training a model on one dataset and then testing it on a different dataset. It is used to assess the generalization ability of models, particularly to evaluate robustness and reliability across different data sources or conditions that were not present in the training data. In this regard, {\em our dataset can be used solely as an evaluation dataset, for testing the robustness of learned representations to state changes and other transformations in real-world scenarios. }

To demonstrate this, we use the PiRO model \cite{PiRO} and our method trained on the ObjectPI \cite{PIE2019} dataset and test it on the proposed ObjectsWithStateChange dataset and show the results in Table \ref{tbl:zero-shot}. Note that between the two datasets, there are 5 seen and 16 unseen categories and all unseen objects. Even for the seen categories, the object transformations and state changes are not present in the training data and the viewpoints are arbitrary. Thus, the training data (from ObjectPI) only has variations in pose where the objects are photographed from pre-defined viewpoints, whereas in the test dataset (from ObjectsWithStateChange), we additionally have state changes and complex transformations of objects photographed from arbitrary viewpoints. From Table \ref{tbl:zero-shot}, we observe that our curriculum learning slightly improves performance on object-level tasks.

Here, we compared the performance of two methods trained on the same ObjectPI dataset and tested on our proposed dataset. Using the same setup, it is also possible to compare the performance of models trained on different datasets to assess how robust and generalizable models trained on different datasets are to state changes. 

\section{Qualitative Fine-grained Retrieval Results}
\begin{figure*}
    \centering
    \includegraphics[width=\textwidth]{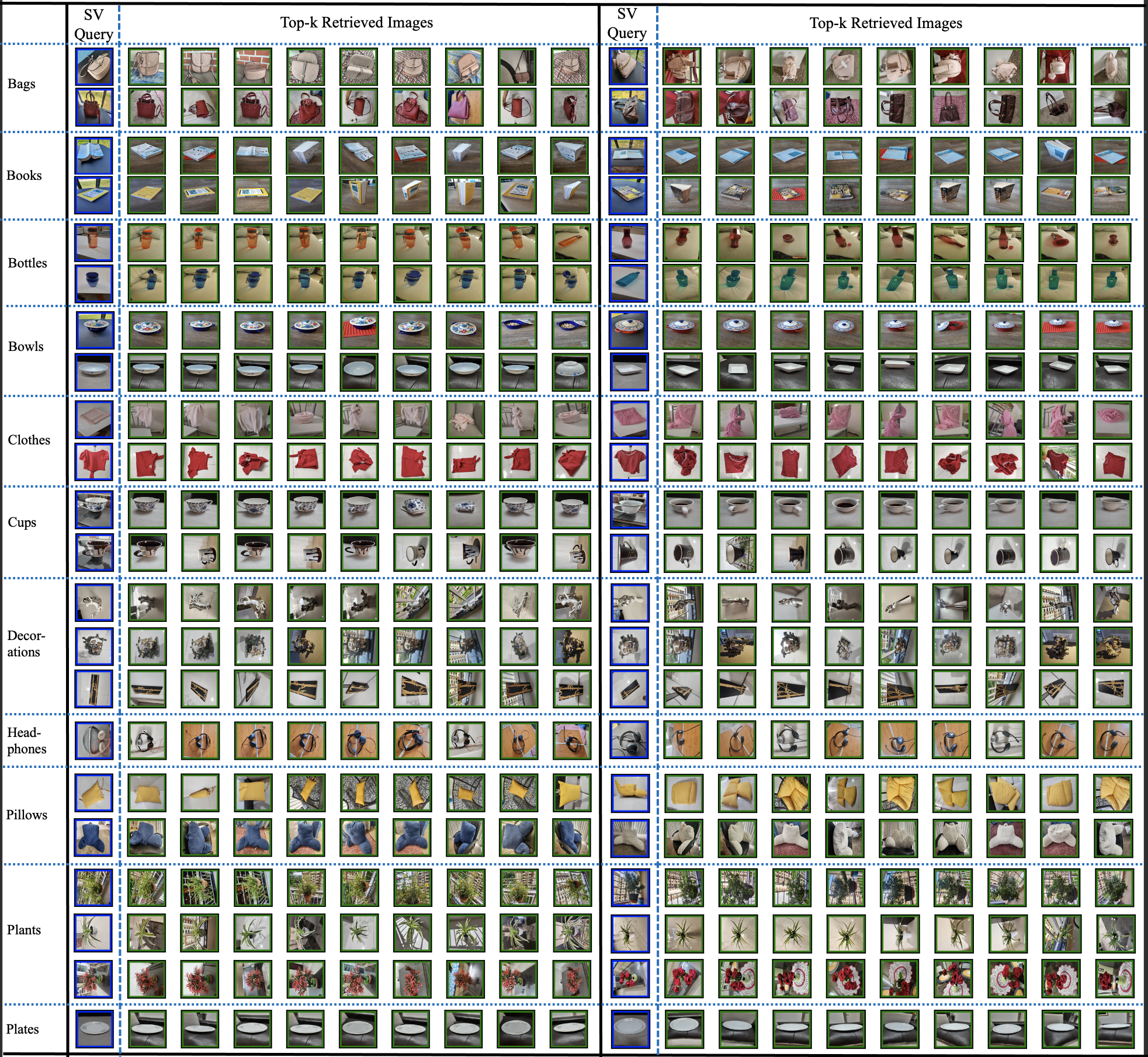}
    \caption{This figure shows single-view object retrieval results of the objects from the first 11 categories. The two columns separated by the thick black line in the middle show results for similar-looking objects within each category. Specifically, in the left column, we show the top-k retrieval results for a SV query image while in the same row of the right column, we show the SV retrieval results for another similar object from the same category. The figure is best viewed when zoomed in. The blue bounding boxes around the images indicate the SV query image while the green and red bounding boxes indicate the correctly and incorrectly retrieved images of the same object-identity as the query image respectively. The results for the other categories are shown on the next page. }
    \label{fig:SV_RETR1}
\end{figure*}

\begin{figure*}
    \centering
    \includegraphics[width=\textwidth]{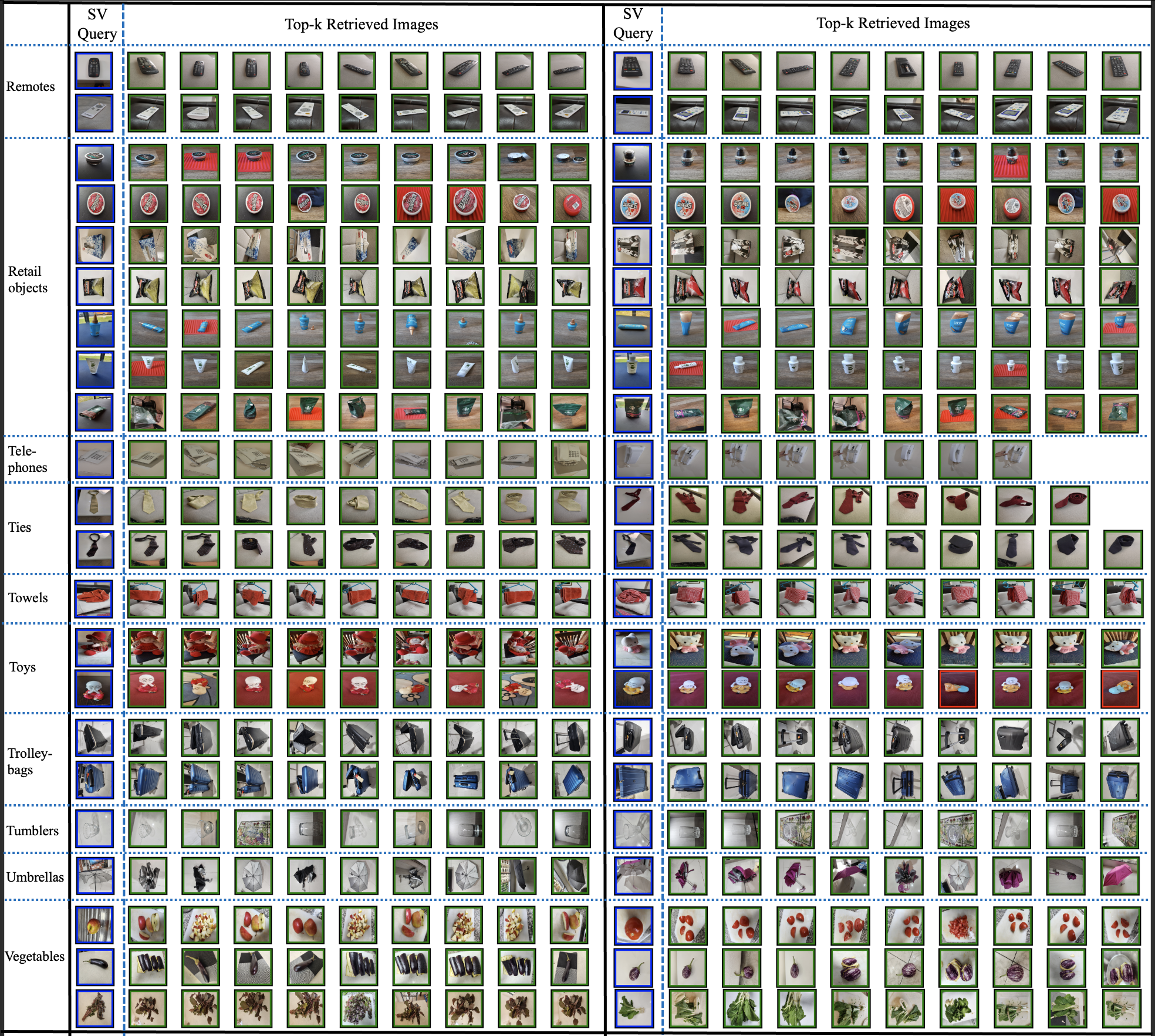}
    \caption{This figure shows single-view object retrieval results of the objects from the remaining 10 categories. Following the same format as the previous figure, the two columns separated by the thick black line in the middle show results for similar-looking objects within each category. Specifically, in the left column, we show the top-k retrieval results for a SV query image while in the same row of the right column, we show the SV retrieval results for another similar object from the same category. The figure is best viewed when zoomed in. The blue bounding boxes around the images indicate the SV query image while the green and red bounding boxes indicate the correctly and incorrectly retrieved images of the same object-identity as the query image respectively. }
    \label{fig:SV_RETR2}

    \end{figure*}

As mentioned earlier in Section 3, our dataset comprises visually similar objects from each of the 21 categories. In this section, we show some qualitative single-image retrieval results of visually similar objects from each category in our dataset in Figs. \ref{fig:SV_RETR1} and \ref{fig:SV_RETR2}.

In these figures, the results are presented in two columns that show the results for similar-looking objects from the same category in each row. Firstly, this illustrates that our dataset has several similar-looking objects from each category with very subtle differences in appearance and hence can facilitate research in fine-grained object retrieval tasks. Secondly, the results indicate that our algorithm is able to retrieve images of the same object-identity with high mAP despite having similar-looking objects from each category and different state changes and other transformations from arbitrary viewpoints. 

\section{Discussion}
\noindent
{\bf Broader Impact and Future Research:} Firstly, we believe that to the best of our knowledge, we are the first to introduce a dataset that captures complex state changes of objects that can be used for category-level and fine-grained object-level recognition and retrieval tasks. This could benefit several real-world applications such as automatic checkout systems \cite{CheckSoft, peopcentsys, RPC2019}, robotic systems, that deal with object state changes upon interaction. Secondly, this dataset can be used for cross-dataset evaluation of methods trained on different datasets and tested for their robustness to state changes and other intricate transformations captured in our ObjectsWithStateChange dataset. Lastly, we provide text annotations describing the visual characteristics of objects that can be used for training multi-modal models such as \cite{CLIP, 4M}. This can be used for adapting such models on object image collections for learning transformation-invariant multi-modal representations and generalized zero-shot evaluation of algorithms as well as visual question answering, which we aim to pursue in the future. 

\noindent
{\bf Challenges:} The following aspects make our dataset challenging: (a) significant appearance changes due to {\em state changes} and other transformations such as {\em arbitrary pose, viewpoint}, lighting, and background changes (b) presence of several similar-looking objects with {\em fine-grained differences} which makes recognition and retrieval of the correct object-identity challenging especially when there are state changes 
(c) images of the objects are captured {\em in the wild} with cluttered and complex backgrounds which facilitates the evaluation of algorithms in real-world scenarios. 

\ifCLASSOPTIONcaptionsoff
  \newpage
\fi

{
    \small
    \bibliographystyle{IEEEtran}
    \bibliography{refs}
}

\end{document}